\documentclass[12pt]{article}
\usepackage{amsmath, amsthm, amssymb, fullpage, setspace}
\title {Evolvability need not imply learnability}
\author{Nisheeth Srivastava}

\newcommand{\R}{\mathcal{R}}
\newcommand{\C}{\mathcal{C}}
\newcommand{\X}{\mathcal{X}}

\theoremstyle{plain}
\newtheorem{defn}{Definition}
\newtheorem{lemm}{Lemma}
\begin{document}
\maketitle

\begin{abstract}
We show that Boolean functions expressible as monotone disjunctive normal forms are PAC-evolvable under a uniform distribution on the Boolean cube if the hypothesis size is allowed to remain fixed. We further show that this result is insufficient to prove the PAC-learnability of monotone Boolean functions, thereby demonstrating a counter-example to a recent claim to the contrary. We further discuss scenarios wherein evolvability and learnability will coincide as well as scenarios under which they differ. The implications of the latter case on the prospects of learning in complex hypothesis spaces is briefly examined. 
\end{abstract}

\section {Introduction}
\label{sec:intro}
Valiant \cite{val08} has recently conjectured that the biological concept of evolvability can be mathematically expressed as a constrained form of computational learning. This presents an opportunity to construct a principled description of the complexity of mechanisms used in biological evolution. Feldman \cite{fel08} has explored this idea further to formalize the equivalence of evolvability to correlational statistical query learning in the PAC framework. The question of finding important classes of functions that are evolvable under this definition has been posed as an open problem to the learning community \cite{felval08}. In this article, we show that monotone disjunctive normal forms (DNFs) are PAC-evolvable under the uniform distribution.  We further show that notwithstanding the claim in \cite{val08} regarding the implication of learnability from evolvability,  this result is insufficient in proving the PAC-learnability of monotone DNFs. We thus draw a distinction between evolvability and learnability which we contend is important in the context of the use of learning algorithms for making predictions on complex hypothesis spaces. Finally, we show that imposing the same conditions on DNF clauses as imposed in recent work by \cite{jacser08} recovers their positive result regarding the PAC-learnability of monotone DNFs.

\section {Background and related work}
\label{sec:back}
PAC-learnability \cite{val84} identifies the prospect of efficiently learning a class of functions used as representations of an observable system's behavior. Specifically,
\begin{defn}
A class of functions $\mathcal{C}$ is said to be $(n,\epsilon,\delta)$ PAC learnable if, given that there exists an optimal hypothesis $f^{*}\in\mathcal{C}$, there exists an algorithm \cal{A} that takes at most poly$(n, 1/\epsilon)$ samples and runs in at most poly$(n, 1/\epsilon)$ time to output with a probability greater than $1 - \delta$ a hypothesis $f\in\mathcal{C}$ that is within an $\epsilon$ of $f^{*}$ in a sense defined by \cal{A}.
\end{defn}
Recall that a PAC model that imposes constraints on the form of the output hypothesis is known as a proper PAC model while one that does not is known as a PAC prediction algorithm. This distinction, while not important for our discussion on evolvability, will have some bearing on our subsequent discussion on the implications of evolvability upon the learnability of DNFs.

Valiant suggests that biological evolution through mutation under natural selection pressure can be modeled as a constrained form of computational learning, where the genome of an organism represents its predictive hypothesis with respect to the environment, and its statistical performance compared to the ideal genome, captured by a correlational measure,  corresponds to both its survival fitness as well as predictive ability. Several objections may be raised here on various biological and methodological grounds, e.g., on the existence of a fixed ideal genome, on the use of correlation as opposed to other measures, etc. However, we concentrate on operating within the existing framework, regarding it to be a plausible model for the biological process.

\textbf{Evolutionary algorithms:} We follow the notation introduced in \cite{felval08} in the rest of this article. Represent the space of attributes $\X = \{0,1\}^n$, a set $\C$ of functions of the attributes, a set $\R$ of representations of functions \footnote{The distinction between $\C$ and $\R$ is best motivated in the context of a biological example: where $\C$ may be considered to be the set of possible phenotypes, $\R$ may be considered to be the set of possible genotypes. This distinction is crucial from the biological point of view, but not essential for any of the mathematical arguments we use here.} and a distribution $D$ over the set $\X$. 

The fitness measure used by an evolutionary algorithm to compare the \textit{performance} of a candidate $r\in\R$ against the ideal function $f\in\C$ is expressed as $Perf_f(r,D) = E_D[r(x)\cdot f(x)]$, where $D$ in our case will be the uniform distribution over $\X$. In practice, the performance will be estimated from a finite number of examples. The empirical performance $Perf_f(r,D,s)$ over $s$ samples is defined as the expectation over $D$ of the random variable $\frac{1}{s}\sum_i{r(x_i)\cdot f(x_i)}$, where $x_1,\cdots,x_i$ are chosen independently from $\X$ according to $D$. 

Following \cite{felval08}, evolutionary algorithms are characterized by a 4-tuple $A = (\R,Neigh,\mu,t)$, where,
\begin{itemize}
 \item{$\R$ is the set of representation functions as defined above.}
 \item {$Neigh(r,\epsilon), r\in R$ is the neighborhood of the representation $r$ and corresponds to the set of representations into which $r$ can randomly mutate. Two further assumptions are made regarding $Neigh(r,\epsilon)$. Firstly, $r\in Neigh(r,\epsilon)$ for all $r$ and $\epsilon$. Secondly\footnote{The latter is clearly an unrealistic assumption and severely weakens the model epistemologically. We briefly discuss the biological implications of this claim later.}, $|Neigh(r,\epsilon)| \leq \text{poly}(n,1/\epsilon)$.}
 \item {$\mu(r,r_1,\epsilon)$ represents the transition probability between $r$ and $r_1\in Neigh(r,\epsilon)$.}
 \item {$t(\epsilon)$ represents a `tolerance' threshold; a minimum quantum of change in fitness for the causative mutation to be considered non-neutral (beneficial/deleterious). t is lower bounded by $\text{poly}(1/n,\epsilon)$.}
\end{itemize}
Lastly, $Neigh, \mu$ and $t$ are assumed to be computable in poly$(n,1/\epsilon).$ At every stage of the algorithm, representative functions corresponding to beneficial mutations (neutral mutations if none beneficial) are chosen to populate the next generation. Thus, the evolutionary algorithm outputs a sequence of beneficial representative functions.   

\textbf{Evolvability:} A mathematical definition of evolvability in the PAC setting emerges naturally by extending the notion of PAC-learning to the domain of evolutionary algorithms. Informally stated, it requires convergence with high probability of the evolutionary algorithm to any ideal fitness function representative $f\in\C$ in a polynomially bounded number of generations $g(n,1/\epsilon)$ of at most polynomially bounded populations $s(n,1/\epsilon).$ More formally,

\begin{defn}
A function class $\C$ is said to be $(n,\epsilon,\delta)$-evolvable over distribution $D$ if there exists an evolutionary algorithm A and polynomials $g(n,1/\epsilon), s(n,/\epsilon)$, for every $f\in\C$, every $\epsilon > 0$ and for all initializations of representative functions $r_0\in\R$,  such that, with probability at least $(1 - \delta)$, there is some sequence of representative functions $r_1,r_2,\cdots$ where $Perf_f(r_{g(n,1/\epsilon)}, D, s(n,1/\epsilon)) > 1 - \epsilon$.
\end{defn}

In \cite{fel08}, Feldman shows that evolvability, as defined above, is precisely equivalent to PAC learning with correlational statistical queries, i.e., queries regarding the correlation of the hypothesis function $r$ with the unknown target function $f$. Since this is a restriction on standard SQ learning, as defined in \cite{kea93}, this observation implies that every concept class known to be SQ learnable is evolvable if the distribution over the domain is assumed to be fixed. Valiant further suggests \cite[Proposition 3.2]{val08} that evolvability of concept class $\C$ over $D$ implies that $\C$ is SQ learnable over $D$. 

The first direction of this argument, namely SQ learnability to evolvability, is correct, and is used in \cite{val08} to demonstrate that monotone conjunctions are evolvable under the uniform distribution. Further, it is also shown that since parity functions are not SQ learnable, they are also not evolvable. We examine the validity of the converse argument later in Section \ref{sec:evolearn} and establish that this need not hold in all cases.
 
\section {Evolving disjunctive normal forms}
\label{sec:evodnf}
We now consider the evolvability of a more general class of Boolean functions, viz. functions expressible as monotone disjunctions of a constant (known) number of conjunctions. The class of monotone DNFs immediately represents a significantly larger space of representations than the set of monotone conjunctions that has so far been shown to be evolvable under the uniform distribution. Intuitively, the set of monotone conjunctions cannot represent hypotheses containing redundancy, multi-functionality and parallelized or modular behavior. Such behavior, however, is strongly characteristic to most complex systems; certainly those in biological domains that are typically considered in the learning community, e.g., genetic interaction networks. DNFs allow such representations in an interpretable manner if the domain's attributes can be mapped in a meaningful manner to the Boolean cube. Therefore, there is a strong incentive to attempt to extend the evolvability framework to the case of monotone DNFs. We see that, up to some semantic ambiguity, this is not a very difficult task to accomplish. 

Broadly speaking, we posit that demonstrating the PAC-evolvability of a monotone k-DNF (disjunction of $k$ conjunctions) is equivalent to  demonstrating the PAC-evolvability of a subset of the $k$ conjunctions separately. In other words, evolving a $k$ term DNF formula iteratively is equivalent to evolving each of the $k$ terms of the DNF individually and then combining the evolved conjuctive clauses to generate the final DNF formula. This immediately requires a generalization of the performance measure $Perf_f(r,D)$ defined in Section \ref{sec:back}. To realize why this must be the case, note that the hypothesis $x_1 \cup x_2 \cup x_3$ will show a high correlation using the existing performance measure with the target function $(x_1\cap x_4\cap x_5) \cup (x_2\cap x_4\cap x_6) \cup (x_3\cap x_7\cap x_8)$, whereas it is nowhere close to the target function in terms of its desired structure or properties. Considering a generalized performance measure that considers the correlation of \textit{individual} conjunctive components of the hypotheses with those in the target solves this problem. We now detail the construction of this measure.  

\textbf{Generalized performance measures:} We construct our generalized performance measure using the performance of the $j^{th}$ conjunction $r_j$ in the hypothesis function $r$ compared to the $i^{th}$ conjunction $f_i$ in the target function $f$. This can be expressed as $$Perf_{f_i}(r_j, D), \quad i, j \in\mathbb{Z}\cap[1,k].$$ We can obtain the performance values for all $i$ and $j$ in $\text{poly}(k)\cdot\text{poly}(n,1/\epsilon)$ time, since each of the individual values is obtainable under the PAC setting. The generalized measure is evaluated as an aggregate of these constituents. Since each $(i,j)$-constituent of the generalized performance measure corresponds to a correlational query, its performance is still meaningful under the CSQ model. 

It is important to note that the method of aggregating the constituent performance measures will affect the interpretation of evolvability of the function class. In the case of DNFs in particular, it should come as no surprise that there are multiple means of defining evolvability. Since approximating one term of a DNF formula will result in a functionally similiar response to learning several, an ambiguity emerges in trying to define what we mean precisely by \emph{evolvability}.

\textbf{Different forms of evolvability}: In general, we can formulate the aggregation of $(i,j)$-constituents in a variety of ways. Simple examples would include considering the maximum, minimum, mean, median or some other statistical combination of each individual performance measure. 
All these generalized performance measures are easily seen to be computable in at most $poly(k)$ iterations of the time complexity of the term-wise CSQ computations for the constituent monotone conjunctions which were earlier shown to have a computational complexity of $poly(n,1/\epsilon)$ individually. Thus, since monotone conjunctions have been shown to be PAC-evolvable, monotone DNFs with a known number of conjunctive clauses are also seen to be evolvable under specific assumptions on the form of the performance objective. We can therefore state that,
\begin{lemm}
The class of monotone Boolean functions is proper PAC-evolvable over the uniform distribution in poly$(k,n,1/\epsilon)$ iterations, where $k$ is the number of conjunctive terms in the target Boolean function and $n$ the number of attributes (literals) in the hypothesis space. 
\end{lemm}
\section {Evolvability and learnability}
\label{sec:evolearn}
It has been earlier suggested that evolvability of a class of functions should imply its learnability \cite[Lemma 2.1]{val08}. However, upon closer examination, we have seen in Section \ref{sec:evodnf} that evolving DNFs can have multiple interpretations, all of which require the use of specific performance measures. The flexibility that we thus create in the definition of performance measures necessitates the negation of the aforegoing statement. In this section, we show how evolvability is not sufficient to prove learnability in the PAC model. Furthermore, we also emphasize how the choice of performance measure afforded in the evolvability framework helps us in evaluating the applicability of various learning algorithms in problem domains where the target functions are DNFs.  We take up this latter consideration first. 

Informally, it may be seen that finding a DNF that while learning theory and its applications seek to model and learn a DNF that is \emph{structurally} similar to the target function in Nature, genotypes and complex systems in general are selected for or against based almost entirely on \emph{functional performance}. Thus, for example, should a particular conjunctive clause in a complicated DNF mapping of an organism's genotype were to be highly correlated with a conjunctive clause in the ideal target function, it shall be selected for with a high probability. To take an even simpler example, it is sufficient for a computer scientist to know well some specialized stream of knowledge in order to be of value to the larger community. This mode of selection, which we here call \textit{functional} appears to be a strong and natural heuristic in the evolution of most natural complex systems. In sharp juxtaposition, modelers in complex systems design and planners in operations research are at times concerned with modeling for `least regret', i.e., ensuring that all possible sequences of outcomes meet a certain performance threshold. This mode of selection, which we here call \textit{structural}, appears to be more an artifact in man-made complex systems. The two paradigms of evolution have previously been addressed in general terms by Dennett \cite{den96}, where the \emph{functional} paradigm is called the `need-to-know' approach and the \emph{structural} paradigm is called the `commando team' approach. In our context, the difference may be enunciated more clearly through the following Lemmas.

\begin{lemm}
It is possible to evolve a disjunctive normal form Boolean function that is \emph{structurally} similar to the target function by optimizing the performance measure $$GenPerf_{f_{DNF}}(r_{DNF}, D) = \min_{i,j = 1\cdots t}{Perf_{f_i}(r_j,D)}$$.
\label{lem:struct}
\end{lemm}

\begin{lemm}
It is possible to evolve a disjunctive normal form Boolean function that is \emph {functionally} similar to the target function by optimizing the performance measure $$GenPerf_{f_{DNF}}(r_{DNF}, D) = \max_{i,j = 1\cdots t}{Perf_{f_i}(r_j,D)}$$. 
\label{lem:func}
\end{lemm}

Where does this distinction become salient? Consider, for example, an association rule mining task. Here, the task is to find a DNF that is structurally similar to some ideal target function that would take the form of disjunctions of all salient association rules (expressed in turn as conjunctions). In this case, the objective function defined in Lemma \ref{lem:struct} would be the appropriate choice for any CSQ PAC algorithm. On the other hand, consider the case of finding gene expression patterns in genetic network data. In this case, it is likely that the target DNF has arisen through a process of multi-functional optimization, a process where the value of the fitness function is of far greater significance that its form. In such a scenario, a CSQ PAC learner should prefer the performance measure given in Lemma \ref{lem:func}. In practice, we often find that clustering and rule mining approaches adopted in machine learning make no distinction between such problem domains. The literature has heretofore not examined the ontology of the presumed target functions in the context of their corresponding problem domains.  Here, we suggest that constructing algorithms bearing in mind the distinction we have explicated may result in significant improvements in accuracy and interpretability of the results. In the absence of such interpretational sensitivity, the prospect of extracting meaningful DNF hypotheses from complex network data using learning algorithms appears dim. On the other hand, while using an evolutionary algorithm instead will allow more realistic and accurate hypotheses to be generated, such an approach will lack the semantic interpretability that propositional learning schemes typically offer. 

Another interesting observation that we can make here concerns the link between evolvability and learnability of DNF formulas.  Learnability is often taken to imply finding a DNF formula that is \textit{structurally} isomorphic with the target DNF. Since, from Lemma 1, monotone Boolean functions are evolvable, this should imply that they are learnable using the specific interpretation of a performance measure described in Lemma \ref{lem:struct}. Showing the PAC-learnability of monotone Boolean functions has been an open problem of outstanding pedigree in the learning community for over 20 years. Thus, it is tempting to consider that evolvability, in the sense defined here, might offer a shortcut to resolving the question in the affirmative. Unfortunately, the heuristic we have suggested is not sufficient to imply learnability in its strictest sense. While using the performance measure defined in Lemma \ref{lem:struct} captures several intuitive properties concerning the fitness of DNF hypothesis from the learning perspective, it does not account for the technical problem of redundancy of the component literals in the conjunctive clauses. The evolutionary algorithm will have a bias towards selecting hypotheses with higher redundancy of literals that occur frequently in the target DNF. Thus, for problems restricted to considering DNF formulas with low redundancy between conjunctive clauses, our performance measure might be expected to perform fairly well. Surprisingly, this heuristic observation appears to concur with insights drawn through a more technical construction by \cite{jacser08} who have very recently shown that monotone DNFs over a uniform distribution on the Boolean cube are PAC-learnable should the component conjunctions not have high redundancy. 

\section {Discussion}

We show that evolvability of a function class, in the sense defined in \cite{val08} does not necessarily imply learnability as has been previously suggested. This contradiction arises from differential semantic interpretations of the correlational performance measure used to judge the fitness of various hypotheses with respect to the ideal function. By adopting a more natural and intuitive interpretation than the one proposed in Valiant's original scheme, we are able to show that the class of monotone Boolean functions will be evolvable under the uniform distribution. We are also able to show that multiple interpretations of evolvability may arise based on considerations specific to application domains. Since learning algorithms in the PAC framework universally confirm to merely one of these interpretations (replicating the target function \emph{structurally}), we further observe the possibility of allowing for a degree of domain-specific sensitivity through the use of PAC-evolutionary algorithms that is beyond PAC-learning algorithms. 

Since the acceleration of research in machine learning commencing in the 90s, several algorithms have been developed for making predictions by learning various aspects of complex systems. As these tools become more readily accessible and implementable, there arises the risk of over-simplifying domains and mistakenly choosing to use learning algorithms to make optimal predictions in scenarios that call for subtler approaches. More specifically, there are several domains that require a \emph{functional} match of some hypothesis with the target Boolean function where the learning algorithm is mathematically designed to find a \emph{structural} match or some approximation thereof. As Valiant suggests in \cite{val08}, we might be better served in attempting to replicate evolutionary processes in learning frameworks to understand and make predictions about phenomena that are historically seen to have emerged evolutionarily. Through our work, we explicate this point further mathematically and demonstrate the prospect of using evolutionary algorithms to solve learning problems that required \emph{satisficing} as opposed to optimal solutions.

\end{document}